\documentclass[lettersize,journal]{IEEEtran}
\usepackage{amsmath,amsfonts}
\usepackage{algorithmic}
\usepackage{algorithm}
\usepackage{array}
\usepackage[caption=false,font=normalsize,labelfont=sf,textfont=sf]{subfig}
\usepackage{textcomp}
\usepackage{stfloats}
\usepackage{url}
\usepackage{verbatim}
\usepackage{graphicx}
\usepackage{cite}
\begin{document}

\title{RAG-KG-IL: A Multi-Agent Hybrid Framework for Reducing Hallucinations and Enhancing LLM Reasoning through RAG and Incremental Knowledge Graph Learning Integration}

\author{Hong Qing Yu, University of Derby and Frank McQuade, Bloc Digital}


\maketitle

\begin{abstract}
This paper presents RAG-KG-IL, a novel multi-agent hybrid framework designed to enhance the reasoning capabilities of Large Language Models (LLMs) by integrating Retrieval-Augmented Generation (RAG) and Knowledge Graphs (KGs) with an Incremental Learning (IL) approach. Despite recent advancements, LLMs still face significant challenges in reasoning with structured data, handling dynamic knowledge evolution, and mitigating hallucinations, particularly in mission-critical domains. Our proposed RAG-KG-IL framework addresses these limitations by employing a multi-agent architecture that enables continuous knowledge updates, integrates structured knowledge, and incorporates autonomous agents for enhanced explainability and reasoning. The framework utilizes RAG to ensure the generated responses are grounded in verifiable information, while KGs provide structured domain knowledge for improved consistency and depth of understanding. The Incremental Learning approach allows for dynamic updates to the knowledge base without full retraining, significantly reducing computational overhead and improving the model's adaptability. We evaluate the framework using real-world case studies involving health-related queries, comparing it to state-of-the-art models like GPT-4o and a RAG-only baseline. Experimental results demonstrate that our approach significantly reduces hallucination rates and improves answer completeness and reasoning accuracy. The results underscore the potential of combining RAG, KGs, and multi-agent systems to create intelligent, adaptable systems capable of real-time knowledge integration and reasoning in complex domains.
\end{abstract}


\section{Introduction}
\IEEEPARstart{T}he rapid advancement of Large Language Models (LLMs), such as GPT-4o, has brought significant breakthroughs in natural language understanding and generation. These models have become invaluable tools across various domains, including text summarization \cite{refp1}, conversational agents \cite{refp2}, and complex question-answering systems. However, despite their capabilities, current LLM frameworks face substantial challenges when it comes to reasoning with structured knowledge, maintaining dynamic knowledge evolution\cite{ref1}, and ensuring accuracy during interactions \cite{ref2, ref3}. Addressing these issues is critical for expanding the applicability of LLM-based agent systems, especially in mission-critical areas such as healthcare, finance, and law, where factual accuracy, reasoning capability \cite{refp3} and explainability \cite{refp4} are crucial.

A major limitation of LLMs lies in their reliance on unstructured text data. These models excel at generating coherent, human-like responses from large volumes of unstructured information but often fail to utilize structured knowledge effectively, such as that found in Knowledge Graphs (KGs). This shortcoming frequently results in hallucinated answers—responses that, while plausible-sounding, are factually incorrect or misleading \cite{ref4}. Moreover, LLMs generally lack the ability to perform contextual reasoning across multiple domains when tasks require a deep understanding of structured, interconnected data \cite{ref5}. 


Another critical issue is the static nature of LLM knowledge. Once trained, these models cannot easily incorporate new information or autonomously update their knowledge base. Any effort to include additional knowledge typically requires retraining or fine-tuning, which is computationally expensive and impractical for many real-world scenarios. 

In light of these limitations, we aim to address the following key research questions:
\begin{itemize}
\item{How to retrieve high quality and relevant data with knowledge integrated RAG-LLM framework to improve answer accuracy with reduced hallucination and enhance reasoning capabilities?} 
\item{What mechanisms can be introduced to enable continuous learning and autonomous knowledge updates?}
\item{How can a multi-agent framework be leveraged to work together to improve the system’s explainability?}
\end{itemize}

We hypothesize that the novel framework we propose in this paper, a multi-agent and hybrid LLM-KG framework that combines with local incremental knowledge learning process, will address the above key questions. Specifically, we expect that such a system will not only reduce hallucinations in the answers but will also enable continuous knowledge updates and offer better explainability and reasoning tasks in a specific domain.







\section{Related work}
\subsection{Hallucination in Generative AI}
Hallucination in generative artificial intelligence (AI) refers to the generation of content that is fabricated, misleading, or not based on factual data \cite{ref6}. This phenomenon has become a major issue as AI systems such as large language models (LLMs) like ChatGPT are increasingly used in real-world applications, often providing seemingly plausible but inaccurate information\cite{ref7}. This issue has been noted in various surveys across different models and applications \cite{ref8} and has been documented extensively across multiple domains and application areas:

Financial Applications: In the financial domain, hallucinations present significant risks, particularly when generative AI is used for decision-making and analysis. Recent research work stresses the necessity for rigorous quality control, diverse data sources, and transparency to mitigate hallucination risks in financial AI applications \cite{ref9}. The same problem has been recognised in many other domains such as healthcare \cite{ref10}, business \cite{ref11}, legal \cite{ref14} and Natural Language Generation (NLG) tasks \cite{ref12}.

Moreover, hallucination also occurs in multimodal AI models, where the generated text may not align with the visual content provided. For example, in image captioning and visual question answering, hallucinations can lead to misinterpretation or incorrect descriptions of objects or scenes, posing significant concerns regarding the practical deployment of such models \cite{ref13}.

The key challenges with hallucinations include the generation of factually incorrect information that may mislead users, especially in sensitive areas like healthcare, finance, and legal contexts. Additionally, the hallucination problem underscores the difficulty in distinguishing between genuine creativity and erroneous outputs, which hinders the deployment of generative AI systems in high-stakes environments\cite{ref9,ref12}.

The evaluation of hallucination in generative AI involves several strategies. The most common methods include statistical metrics, model-based metrics, and human evaluations \cite{ref12}. Recent research has emphasized the importance of identifying hallucinated content through in-context learning (ICL) settings. Jesson et al. presented a Bayesian model to estimate the probability of hallucination in ICL tasks by calculating response probabilities based on context\cite{ref15}.

Another approach to measuring hallucination includes the use of a cross-encoder model and n-gram overlap metrics to promote more grounded responses \cite{ref16}. These measurements provide an empirical framework to understand hallucination better and evaluate the mitigation approaches.

Addressing hallucination in generative AI has received significant attention. Several strategies have been proposed to mitigate hallucinations, ranging from improvements in training methodologies to more advanced inference techniques.

Retrieval-Augmented Generation (RAG): RAG combines generative models with information retrieval techniques to ensure that the generated output is grounded in real, verifiable data. Béchard et al.\cite{ref17} demonstrated the effectiveness of RAG in reducing hallucinations in structured outputs like workflow generation from natural language instructions.

Channel-Aware Domain-Adaptive Generative Adversarial Network (CADAGAN): Grayson et al.\cite{ref18} proposed CADAGAN, which modifies generative models by introducing channel-aware processing and domain-adaptive learning. This technique helps in mitigating hallucinations by aligning generated output with domain-specific knowledge while maintaining linguistic coherence. In addition, the research shows that self-refining approach is another approach for training an agent incrementally rather than batch process.  

Genetic Algorithm for Grounded Answer Generation (GAuGE): Kulkarni et al.\cite{ref16} introduced a genetic algorithm-based grounded answer generation method to minimize hallucination. This method effectively maintains high relevance by cross-checking with retrieved search engine results and encouraging grounding through a balanced fitness function.

Domain-Specific Adaption and Knowledge Graph Utilization: Techniques such as domain adaptation and knowledge graph integration have shown promise in controlling hallucination. Towhidul Islam Tonmoy et al.\cite{ref19} reviewed multiple mitigation strategies, including knowledge retrieval and domain adaptation to enhance fact consistency in generated texts.

\subsection{RAG-LLM and Knowledge Graph-based LLM Solutions and Their Current Limitations}

The use of Retrieval-Augmented Generation (RAG) to enhance LLMs has gained significant popularity in recent years. RAG aims to address the inherent limitations of LLMs by supplementing them with up-to-date and domain-specific knowledge from external databases. This approach has been successful in reducing hallucinations, improving reasoning capabilities, and enhancing response quality. RAG provides an external source of information, enabling the LLM to generate content that is both timely and factually accurate, rather than relying solely on the model's pre-trained internal knowledge \cite{ref20}.

One of the key benefits of RAG is its ability to mitigate hallucinations in LLMs by providing access to authoritative information during the generation process. For example, the integration of retrieval mechanisms, such as dense vector similarity search, enhances the LLM's ability to answer out-of-scope questions by supplying relevant facts from external sources \cite{ref21}. However, RAG has several limitations. Existing retrieval systems are often constrained by the retrieval mechanism, which may fail to provide adequate granularity for complex queries or may introduce irrelevant information, commonly referred to as "hard negatives," that can mislead the LLM during the generation process \cite{ref22}.

Recent work also highlights that RAG may not be fully effective in improving the deep reasoning capabilities of LLMs. In the study titled "How Much Can RAG Help the Reasoning of LLM?" \cite{ref23} it was found that while RAG could aid the reasoning process by incorporating relevant documents, it falls short when it comes to performing deeper or more complex inferences \cite{ref23}. This is mainly because LLMs are limited in terms of reasoning depth, and incorporating external documents often requiring preprocessing and filtering to remove noise, which may degrade reasoning capabilities rather than enhancing them.

In addition, the RAFT-RAG framework \cite{refp21} emphasizes that data pre-selection is critical for enhancing input quality in domain-specific Retrieval-Augmented Generation (RAG) systems, especially in question-answering tasks. RAFT adapts pre-trained language models to specific domains by incorporating both relevant and distractor documents during training. This helps the model to discern useful information while ignoring irrelevant content, thus improving its robustness and accuracy when handling domain-specific questions. This strategy ensures that models are better prepared for real-world applications, where not all retrieved data is useful, reinforcing the value of pre-selecting high-quality data inputs for domain-focused RAG systems.

The other trend of RAG-LLM development is based on Knowledge Graph-based approaches integrated with RAG to create hybrid models like the KG-RAG \cite{ref21} and KRAGEN frameworks \cite{ref24}, which aim to address the shortcomings of retrieval-based systems by providing structured and accurate domain-specific knowledge. These models construct Knowledge Graphs (KGs) from unstructured text, which then supports question answering and reasoning tasks by offering an explicit representation of entities and relationships. However, these hybrid systems still face challenges such as managing the complexity of integrating KGs with LLMs and the inherent limitations of retrieval mechanisms, which can introduce noise or irrelevant information, affecting overall response quality.  

\subsection{Continual Learning Challenges in LLM and the Importance of Incremental Learning}


Incremental learning (IL), also known as continual learning, provides an alternative approach that allows LLMs to update their knowledge incrementally without complete retraining. The review by Wang et al. \cite{ref25} explores various approaches to incremental learning, including rehearsal methods, parameter-efficient tuning, and meta-learning, which enable models to adapt to new information while mitigating the issue of catastrophic forgetting. Catastrophic forgetting occurs when a model loses previously learned knowledge upon learning new tasks, a major challenge in current LLM architectures.

Incremental learning is particularly useful for real-time applications where new information must be integrated immediately to maintain model relevance. Unlike traditional batch training, IL allows for knowledge updates in a way that resembles human learning—where new information builds on existing knowledge without erasing prior understanding. Techniques such as parameter-efficient finetuning (e.g., LoRA) and dynamic-architecture methods have been employed to facilitate efficient knowledge updates while maintaining the model's previous competencies \cite{ref26}.

In summary, adopting incremental learning methodologies in LLMs would significantly enhance their adaptability and efficiency, reduce computational overhead associated with batch retraining, and allow models to maintain up-to-date knowledge in real-time applications.

\section{Proposed Framework}

\begin{figure*}[h]
\centering
\includegraphics[width=\textwidth]{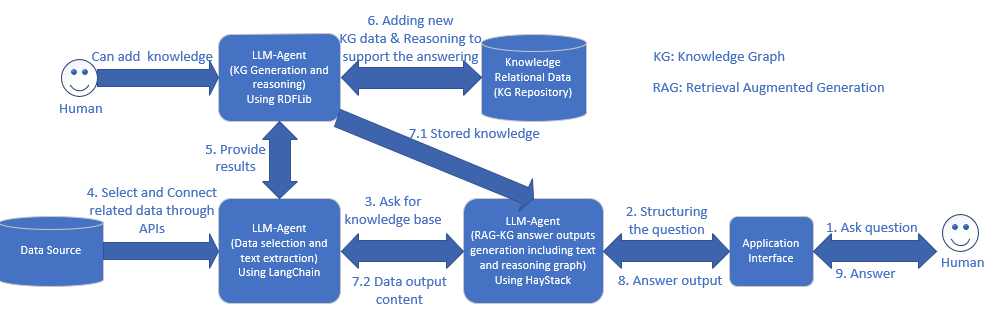}
\caption{The proposed RAG-KG-IL framework}
\label{fig1}
\end{figure*}

Based on the studies of recent work, we present the design and operational structure of our novel framework, a hybrid system that integrates Large Language Models (LLMs) with Knowledge Graphs (KGs) reasoning and an incremental learning approach using a multi-agent \cite{ref27} (LangChain \cite{ref29}) and (Haystack \cite{ref28}) architectures. The framework aims to address the limitations of current LLM architectures by enhancing reasoning capabilities to reduce hallucination and enabling continuous learning with long context data. The key innovation lies in the new methodology of a multi-agent system and training a domain expert agent through a question-and-answer process incrementally by leveraging RAG-Agent and Knowledge Graphs technologies.

Figure \ref{fig1} illustrates the core components and workflow of the proposed framework. In the following subsections, each component is described, detailing how they interact to improve accuracy, efficiency, and reasoning capabilities.

\begin{table*}[ht]
\centering
\begin{tabular}{|>{\raggedright\arraybackslash}p{4cm}|>{\raggedright\arraybackslash}p{12cm}|}
\hline
\textbf{Purpose} & \textbf{Prompts} \\ \hline

KG and text data fusion to generate answer & 

I have the following Knowledge Graph data and generated text data that related to medical conditions learned from the answers provided by the NHS website. \\
& \\ &
Knowledge Graph data: \textless KGData \textgreater \\
 & Text data: \textless TextData \textgreater \\

& \\ & Your task is to take the following query related to a medical condition as an input to extract useful information by parsing and reasoning through the Knowledge Graph triples to perform two tasks: \\

& 1. Validate the text answers from Text data to remove inconsistent answers. \\
& 2. Restructure the related triples as the text output to integrate with the validated Text Data to produce the final answer. \\ & \\

& Query: \textless question \textgreater \\ & 

\\ & "role": "system", "content": prompt, 
\\ & "role": "user", "content": "Provide me the answer."\\

\hline

Data Selection and Extraction & You are a web link search engine responsible for finding the most relevant web links from the provided list based on the user's query. Your process is guided by the following instructions: \\ & \\
& Link Relevance: Ensure that the links selected are directly related to the medical term or condition in the user's query. Include only links that provide precise and useful information. \\ 
& Pre-existing Knowledge: Leverage relevant knowledge from the Knowledge Graph repository, if available, to enhance the quality of link selection and ensure comprehensive coverage.
\\ 
& Requirements: Provide the output as a single comma-separated string containing only URLs from the provided list. Strictly avoid any comments, human language, or further explanations.
\\ 
& Your task is to carefully analyze the user query along with the pre-existing knowledge and identify only the validated web links from the provided list. \\ & \\
& Input Elements: Query: \textless question\textgreater,
\\ & Knowledge Source: \textless kgPath\textgreater, 
\\ & List of Links: \textless listOfWeblink\textgreater. \\ 
& APIs web content query code path: \textless codepath\textgreater, \\ & 
\\ & {"role": "system", "content": "Prompt"}
\\ & {"role": "user", "content": "Here is the query: " +question+ " Can you get all the related web URLs and extract the content from them to save at " + tempPath + "?"}
\\

\hline

\end{tabular}
\caption{Prompts Engineering for data fusion and output graph representation to the interface \label{tab:tablePrompt1}}
\end{table*}

\subsection{Application Interface}

\begin{figure*}[h]
    \centering
    \includegraphics[width=\linewidth]{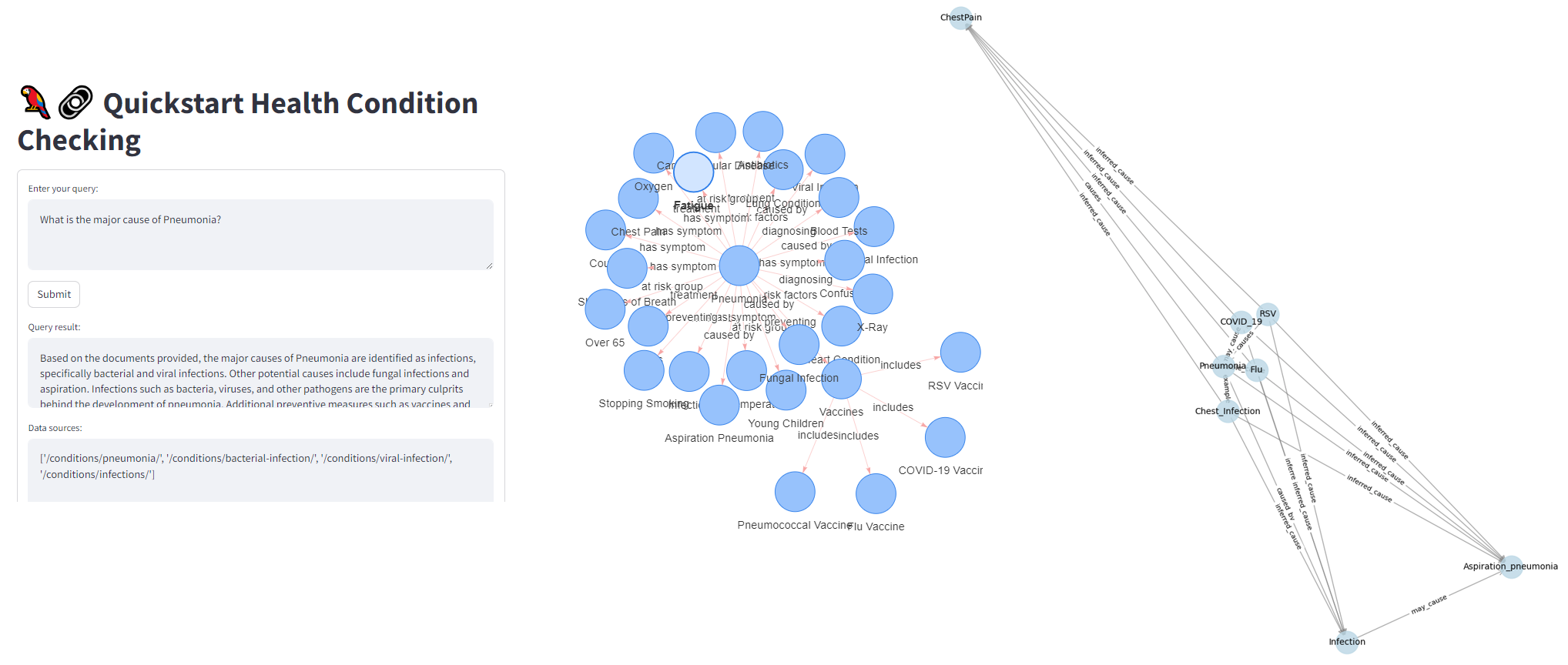}
    \caption{Interface view, left: question and answer including sources, middle: subset of knowledge graph related to the question, right: causality reasoning analysis.}
    \label{fig:interface}
\end{figure*}

The user interaction begins with the Application Interface, where users pose their questions. This interface is designed to ensure ease of use and transparency, converting user queries into a structured format that can be processed by the subsequent agents (Step 1). The Application Interface ensures that users are presented with both structured and graphical outputs, making the final responses easy to interpret (Step 9). Figure \ref{fig:interface} shows an example of question and answering output scenario. You can also see the graph representation output of the linked terms or documents generation prompts in the second row of Table \ref{tab:tablePrompt1}. The reasoning result will be discussed in detail in the evaluation results section.    

\subsection{LLM-Agent (RAG-KG-IL Answer Generation Engine)}

In Step 3 of the RAG-KG-IL workflow, the system generates an answer by fusing retrieved data from selected sources and sub-knowledge graph data relevant to the user's question. The process is divided into three key stages: data fusion, vectorization, and answer generation.


The data fusion and validation LLM Prompt for our domain specific case study scenario is presented in the first row of Table \ref{tab:tablePrompt1}. 

The retrieved text data \( D_{\text{retrieved}} \) and the sub-knowledge graph data \( K_{\text{sub}} \) are first combined into a unified representation \( F_{\text{fused}} \) using a concatenation function:
\[
F_{\text{fused}} = \text{concat}(D_{\text{retrieved}}, K_{\text{sub}})
\]
Here, the \(\text{concat}(\cdot, \cdot)\) function merges the textual and structured data into a coherent intermediate format.

The fused data representation \( F_{\text{fused}} \) is then transformed into a numerical vector representation \( V_{\text{fused}} \) using a vectorization function \( v(\cdot) \):
\[
V_{\text{fused}} = v(F_{\text{fused}})
\]
This step converts the fused data into a machine-readable form, enabling downstream processing by the language model.

The vectorized representation \( V_{\text{fused}} \) is passed to the answer generation function \( g(\cdot) \), which synthesizes the final response:
\[
A(Q) = g(V_{\text{fused}})
\]
Here, \( A(Q) \) represents the generated answer for the user’s query \( Q \). The function \( g(\cdot) \) utilizes a retrieval-augmented generation (RAG) pipeline from Haystack framework \cite{refp22} to integrate seamlessly into the LLM agent to generate accurate from vectorized data, contextually relevant answers. It integrates the most relevant contextual information to produce accurate and contextually aligned answers.





The reasoning process is applied to the fused data \( F_{\text{fused}} \) from Step 3, producing a reasoning graph \( G_{\text{reason}} \). The graph \( G_{\text{reason}} \) is defined as:
\[
G_{\text{reason}} = (V, E) = \text{reasoning\_graph}(F_{\text{fused}})
\]
where:
\begin{itemize}
    \item \( V \): The set of nodes representing key entities, concepts, or terms derived from the retrieved data and sub-knowledge graph.
    \item \( E \): The set of edges representing relationships or causal links inferred between the entities.
\end{itemize}

The generated graph provides:
\begin{enumerate}
    \item \textbf{Causal Pathways}: Displays relationships and reasoning paths between entities, offering insights into how the answer was derived.
    \item \textbf{Supporting Knowledge}: Links the terms and concepts back to the knowledge graph and retrieved data for verification and transparency.
\end{enumerate}

The reasoning graph serves to:
\begin{itemize}
    \item Enhance transparency by illustrating the relationships and logic underlying the answer.
    \item Provide users with a visual, easy-to-understand explanation of the reasoning process.
\end{itemize}

The prompt for the function of graph output generation is presented in the first row of Table \ref{tab:tablePrompt2}. 

\subsection{LLM-Agent (Data Selection and Text Extraction Using LangChain)}

The LLM-agent determines the most relevant data sources by semantically understanding the user’s question \( Q \) and matching it with the indexed data sources. Let \( \text{Sources} \) represent the set of all indexed data sources. The relevant data sources are identified as:
\[
\text{Relevant\_Sources} = \text{LLM\_agent}(Q, \text{Sources})
\]
where \( \text{LLM\_agent}(\cdot, \cdot) \) represents the agent’s capability to dynamically identify semantically relevant sources based on the question \( Q \) using LangChain LLM framework. The data selection and extraction Prompt for our domain specific case study scenario is illustrated in the second row of Table \ref{tab:tablePrompt1}.

Once the relevant sources are selected, the LLM-agent queries these sources to retrieve the necessary data. This is expressed as:
\[
D_{\text{retrieved}} = \bigcup_{D_i \in \text{Relevant\_Sources}} \text{API\_query}(D_i, Q)
\]
where \( \text{API\_query}(\cdot, \cdot) \) denotes the process of querying the data source \( D_i \) with the question \( Q \) to extract relevant information.

The retrieved data \( D_{\text{retrieved}} \) is then passed to subsequent steps:
\begin{itemize}
    \item To Step 7.2 for fusion with sub-knowledge graph data.
    \item To Step 5 for next agent to do knowledge graph generation and reasoning.
\end{itemize}


\subsection{LLM-Agent (KG Generation and Reasoning Using RDFLib)}

A key feature of the proposed framework is the use of the LLM-Agent for KG Generation and Reasoning (Step 6). Using RDFLib reasoning tools, this component adds new knowledge to the Knowledge Graph and performs reasoning to support subsequent answers. The system can enrich its existing knowledge with new relational data and enhance reasoning capabilities, ensuring that it adapts to user interactions over time. The updated KG is stored in the Knowledge Relational Data repository (Step 7.1), making it available for future queries. The graph generation and reasoning prompts are presented in the Table \ref{tab:tablePrompt2}.  

In Step 6, the LLM-agent plays a central role in extracting new knowledge and reasoning over the enriched knowledge graph using RDFLib tools. The LLM-agent extracts entities and relationships from the retrieved data \( D_{\text{retrieved}} \) and the user query \( Q \):
\[
\text{Extracted\_Knowledge} = \text{LLM\_extract}(D_{\text{retrieved}}, Q)
\]
This extracted knowledge is then used to enrich the existing knowledge graph:
\[
K_{\text{KG,new}} = K_{\text{KG,old}} \cup \text{Extracted\_Knowledge}
\]

The enriched knowledge graph \( K_{\text{KG,new}} \) is processed using RDFLib reasoning tools to infer new facts and generate a reasoning graph:
\[
G_{\text{reason}} = \text{reason}(K_{\text{KG,new}})
\]
Here, \( G_{\text{reason}} = (V, E) \), where:
\begin{itemize}
    \item \( V \): The set of nodes representing entities or concepts.
    \item \( E \): The set of edges representing inferred relationships or causal links.
\end{itemize}

\begin{table*}[ht]
\centering
\begin{tabular}{|>{\raggedright\arraybackslash}p{4cm}|>{\raggedright\arraybackslash}p{12cm}|}
\hline
\textbf{Purpose} & \textbf{Prompts} \\ \hline
Graph output generation & 
I have the following text related to a medical condition. First summerise the keywords that present the condition names, symptome names, treatment names, prevention method names, diagnosing method names and who is at rick from the conditions. Then, create graph nodes and edges to represent the relationships between key medical terms that relation maximumly include [disease names, possible cause, symptoms, treatment, diagnosing, Preventing, Who's at risk] in a structured format. The nodes should only represent important medical terms without any descriptive or verb words and key concepts, and the edges should represent the relationships between them.
\\ & Ensure the output follows this Python code format:
\\ & nodes = []
\\ & edges = []
\\ & nodes.append(Node(id="Term1", label="Description1", size=25, shape="circularImage"))
\\ & nodes.append(Node(id="Term2", label="Description2", size=25, shape="circularImage"))
\\ & edges.append(Edge(source="Term1", label="relation", target="Term2")) \\ & 

Text: \textless textAnswer \textgreater \\ & 

\\ & Provide the nodes and edges in a JSON formate that must be validated JSON file. Remember no further comments or explanations are needed to add to the file only the pure JSON data no any comments are needed. \\ &

\\ & "role": "system", "content": prompt, \\ & "role": "user", "content": "Provide me the graph notes.json file" \\ \hline
Graph Generation and update & 
I have the following text related to a medical condition. Create graph nodes and edges to represent the relationships between medical terms and key sentences in a structured format. The nodes should represent important medical terms and key concepts, and the edges should represent the relationships between them. \\ &

\\ & Ensure the output follows this Python code format:
\\ & nodes = []
\\ & edges = []
\\ & nodes.append(Node(id="Term1", label="Description1", size=25, shape="circularImage"))
\\ & nodes.append(Node(id="Term2", label="Description2", size=25, shape="circularImage"))
\\ & edges.append(Edge(source="Term1", label="relation", target="Term2"))
\\ & Text: \textless tempath \textgreater \\ & 
\\ & Provide the nodes and edges in the specified Python code format. \\ 
\\ & {"role": "system", "content": prompt},
\\ & {"role": "user", "content": "Provide me the graph"}\\
\\ & \#Then we will ran our RDFlib code to update the KG repository with this new graph \\
\hline

\end{tabular}
\caption{Prompts Engineering for data selection, extraction, knowledge graph generation and update \label{tab:tablePrompt2}}
\end{table*}

\subsection{Knowledge Graph Repository incrementally growth and Human-in-the-Loop Knowledge Addition}

The Knowledge Graph Repository stores structured data relevant to user queries (Step 7). It is continuously updated as new queries are processed and knowledge is added. This evolving knowledge base allows the system to improve its ability to respond to increasingly complex and detailed queries over time. The Knowledge Graph's relational structure is essential for maintaining accuracy and consistency across responses.

Another key aspect of the proposed framework is the ability for human users to add knowledge directly into the Knowledge Graph. This "human-in-the-loop" element (Step 6) provides an opportunity for domain experts to enhance the knowledge base, ensuring the accuracy of data and expanding the graph's capabilities with curated insights that may not be accessible through automated extraction alone.


\section{Evaluation Methodology}


This section outlines the methodology used to evaluate the accuracy of the answers generated by our proposed RAG-KG-IL-Agent and the GPT-4o model. The evaluation was conducted as a case study using the UK NHS A-Z disease contents and their APIs. The evaluation process went through multiple iterations to address challenges and limitations identified during the initial phase.

We created a set of 20 questions covering 10 interconnected diseases, focusing on aspects such as symptoms, prevention, treatment, and causes. For example, one of the questions is, "What are the symptoms of pneumonia and how can it be prevented?" Ground truth content for these questions was obtained from NHS websites to ensure the reliability of the answers. The evaluation focused on two key aspects: the accuracy of the answers against the ground truth content (excluding clinical accuracy) to compare hallucination rates across the proposed system RAG-KG-IL, the RAG-only system, and the original GPT-4o model, as well as assessing knowledge creation and reasoning capabilities during the incremental learning process. 

\subsection{Knowledge Creation and Causality Reasoning Capabilities Evaluation Method}
We will monitor the growth of the knowledge graph by evaluating the increase in triple size and comparing the number of medical terms and relationships after answering 1, 10, and 20 questions. This approach will help assess the causality reasoning capabilities of the proposed framework and its impact on improving the accuracy and completeness of the generated answers. Graphical representations will be generated to illustrate the significant changes during the incremental study process.

The growth is calculated as:
\[
\Delta K_{\text{triples}} = |K_{\text{after}}| - |K_{\text{before}}|
\]
where:
\begin{itemize}
    \item \( |K_{\text{triples}}| \): Represents the total number of triples in the knowledge graph.
    \item \( K_{\text{after}} \): The knowledge graph state after answering a set of questions.
    \item \( K_{\text{before}} \): The knowledge graph state before answering the questions.
\end{itemize}

For evaluating causality reasoning capabilities, the system monitors the increase in the number of relationships (\( E \)) in the reasoning graph:
\[
\Delta G_E = |E_{\text{after}}| - |E_{\text{before}}|
\]
where:
\begin{itemize}
    \item \( |E| \): Represents the total number of edges (relationships or causal links) in the reasoning graph.
    \item \( E_{\text{after}} \): The set of relationships in the reasoning graph after reasoning is applied.
    \item \( E_{\text{before}} \): The set of relationships in the reasoning graph before reasoning is applied.
\end{itemize}

These metrics provide insights into the framework’s ability to dynamically expand its knowledge base and enhance its reasoning capabilities as more queries are processed.

\subsection{Scalability of time cost Evaluation Method}
During the evaluation, we will do a limited scalability evaluation to record the time required to answer questions and retrieve knowledge as the size of the knowledge graph increases from 1 question to 20 questions. The results will help determine whether the system scalability is significantly impacted by the growth of the knowledge graph triples. 


The total time cost is calculated as:
\[
T_{\text{total}} = T(L_i) + T(R_i) + T(A_i)
\]
where:
\begin{itemize}
    \item \( T_{\text{total}} \): The total time cost for answering a question.
    \item \( T(L_i) + T(R_i) + T(A_i)\): The time required to answer the \( i \)-th question, which includes LLM-agent API response time, knowledge graph reasoning, and answer generation.
\end{itemize}

This evaluation helps assess how the increasing size of the knowledge graph impacts the system’s performance. While this analysis is limited to 20 questions due to testing constraints such as the LLM API usages, machine capability and research resources, it provides valuable insights into the framework’s scalability and efficiency as the knowledge base grows.

\subsection{Initial Design of Accuracy Evaluation Method}

The initial evaluation design was aimed to create an automated scoring agent to assess the accuracy and completeness of the answers generated by the RAG-KG-IL Agent, RAG only and GPT-4o. The evaluation process consisted of using a scoring agent to compare the generated answer against the ground truth context, focusing on three key aspects of accuracy, completeness and inaccuracy which are described in the prompt given in Table \ref{tab:evaluation_results}.  

\begin{table*}[h]
    \centering
    \begin{tabular}{|p{0.45\textwidth}|p{0.45\textwidth}|}
        \hline
        \textbf{Prompt} & \textbf{Evaluation Results} \\
        \hline
        \begin{minipage}[t]{0.95\linewidth}
            You are an expert evaluator tasked with assessing the accuracy of an answer provided by an AI system. Your role is to evaluate the answer based on three criteria: accuracy, completeness, and the presence of incorrect information, using the ground truth provided. Your specific tasks are as follows:
            \begin{itemize}
                \item \textbf{Accuracy (30 points)}: Determine if the statements in the answer are properly referenced from the ground truth. Each statement must be traceable to the provided information.
                \item \textbf{Completeness (30 points)}: Evaluate whether the answer includes all the relevant key information present in the ground truth.
                \item \textbf{Inaccuracy Penalty (40 points)}: Identify any incorrect or unsupported statements. Each such statement will incur a significant penalty of up to 40 points.
            \end{itemize}
            Carefully read the question, ground truth information, and the given answer. Identify any sentences in the answer that contain information not validated or incorrect according to the ground truth. Output these invalidated or incorrect sentences, each separated by "::".
            \begin{itemize}
                \item \textbf{Question}: Question text
                \item \textbf{Ground truth}: Relevant information from ground truth
                \item \textbf{Answer}: Generated answer to evaluate
            \end{itemize}
        \end{minipage} &
        \begin{minipage}[t]{0.95\linewidth}
            \textbf{Acne: What are the causes of Acne and its treatment?} \\
            \textbf{RAG-KG-IL Answer Score: 95} \\
            The answer is accurate (28/30) as it correctly identifies causes and treatments for acne based on ground truth. It covers key points (27/30) like hormonal changes, bacteria, and treatment options. There are minimal inaccuracies or misleading information (38/40), missing only minor aspects such as the document folder location. Overall, the answer is informative and well-aligned with the ground truth. \\\\
            \textbf{GPT-4o Answer Score: 80} \\
            The answer is correct (25/30) with coverage of causes like hormonal changes, sebum overproduction, and genetic predisposition. It lacks a few treatment specifics (e.g., mild self-help guidance) for completeness (25/30). Minor inaccuracies include overstated roles of diet and stress, with peripheral treatments not mentioned in the ground truth, leading to moderate penalties (30/40). \\\\
            \textbf{Pneumonia: What are the causes of Pneumonia and its treatment?} \\
            \textbf{RAG-KG-IL Answer Score: 95} \\
            The answer is mostly correct regarding causes and treatments (27/30). It is largely complete, covering infections and treatments, though preventive measures are additional details not in the ground truth (25/30). There's no penalty for inaccurate information (38/40). Overall, the answer is accurate and relevant, with a minor penalty for unsupported details. \\\\
            \textbf{GPT-4o Answer Score: 65} \\
            The answer is largely accurate about causes (27/30) and covers treatment options well (27/30), capturing key elements from the ground truth. A slight penalty (11/40) is applied for mentioning unverified specifics, like mechanical ventilation and comprehensive preventive measures, which are not detailed in the ground truth. Overall, it effectively conveys the necessary information. \\
        \end{minipage} \\
        \hline
    \end{tabular}
    \caption{Prompt and evaluation results for answers generated by RAG-KG-IL Agent and GPT-4o using initial agent evaluation design}
    \label{tab:evaluation_results}
\end{table*}

The scoring was designed to generate a numerical score by combining accuracy and completeness measures. However, the initial implementation revealed significant issues with consistency and correctness. From Table \ref{tab:evaluation_results}, we can clearly see that the sum of the scores are not correct. For example, Acne's RAG-KG-IL answer scoring was 28+27+38 $\neq$ 95 and Pneumonia's RAG-KG-IL answer scoring was 27+25+38 $\neq$ 95.

In addition, the LLM-based scoring agent often failed to provide consistent evaluations. For example, "A slight penalty (11/40) is applied for mentioning unverified specifics", however reducing 29 marks is not a slight penalty. By contrast, for the "Minor inaccuracies include overstated roles of diet and stress, with peripheral treatments not mentioned in the ground truth, leading to moderate penalties (30/40).", only 10 marks were reduced for moderate penalties. Due to these significant issues, it became evident that relying solely on an automated agent scoring approach for such nuanced evaluation tasks was insufficient. This initial test of the evaluation also can be taken as an evidence to show the hallucination and low capability of reasoning problems of current LLM systems. 

\subsection{Modified Design of Accuracy Evaluation Method}

To address these challenges, a modified evaluation method was developed that focused on two primary components: \textbf{accuracy matching} and \textbf{human-verified completeness}. The updated approach involved the following steps:

\begin{enumerate}
\item \textbf{Accuracy Matching (Truth Checking Agent)}: Instead of scoring the answer, a truth-checking agent was employed to identify and list all invalidated words or statements in the generated answer, comparing them against the ground truth. The truth-checking agent will flag any inaccurate or unverified content, providing a list of discrepancies. This approach eliminated the need for complex scoring calculations, focusing instead on identifying errors.
\[
\text{Errors}_{\text{invalid}} = \{w_i \in A(Q) \ | \ w_i \notin \text{Ground\_Truth}(Q)\}
\]
where:
\begin{itemize}
    \item \( \text{Errors}_{\text{invalid}} \): The set of invalidated words or statements identified by the truth-checking agent.
    \item \( w_i \): Individual words, phrases, or statements in the generated answer \( A(Q) \).
    \item \( \text{Ground\_Truth}(Q) \): The set of all valid components derived from the ground truth context for the question \( Q \).
\end{itemize}

The total number of invalid items is calculated as:
\[
\text{Count}_{\text{invalid}} = |\text{Errors}_{\text{invalid}}|
\]

\item \textbf{Human-Verified Completeness}: Following the accuracy matching step, a human evaluator reviewed the completeness of the answer. This involved counting the number of discrepancies identified by the truth-checking agent and manually verifying whether the generated answer comprehensively addressed the question according to all related data displayed in the ground truth documents. 
\[
\text{Missing\_Items} = \{x \in \text{Ground\_Truth}(Q) \ | \ x \notin A(Q)\}
\]
where:
\begin{itemize}
    \item \( \text{Ground\_Truth}(Q) \): The set of all valid items (e.g., facts, symptoms, treatments) relevant to the question \( Q \) from the ground truth content.
    \item \( A(Q) \): The set of items included in the generated answer for the question \( Q \).
    \item \( x \): An individual item from the ground truth.
    \item \( \notin \): Denotes that \( x \) is not addressed in the generated answer \( A(Q) \).
\end{itemize}

\end{enumerate}

Both the Accuracy Matching and Human-Verified Completeness evaluations were conducted for the RAG-KG-IL Agent, RAG only Agent and GPT-4o answers, with a comparison of scores between the two to assess their relative performance.

By incorporating human verification into the evaluation process, the modified design ensured that the results were both reliable and robust. 


\section{Evaluation Results}

\begin{figure*}[h]
    \centering
    \includegraphics[width=\linewidth]{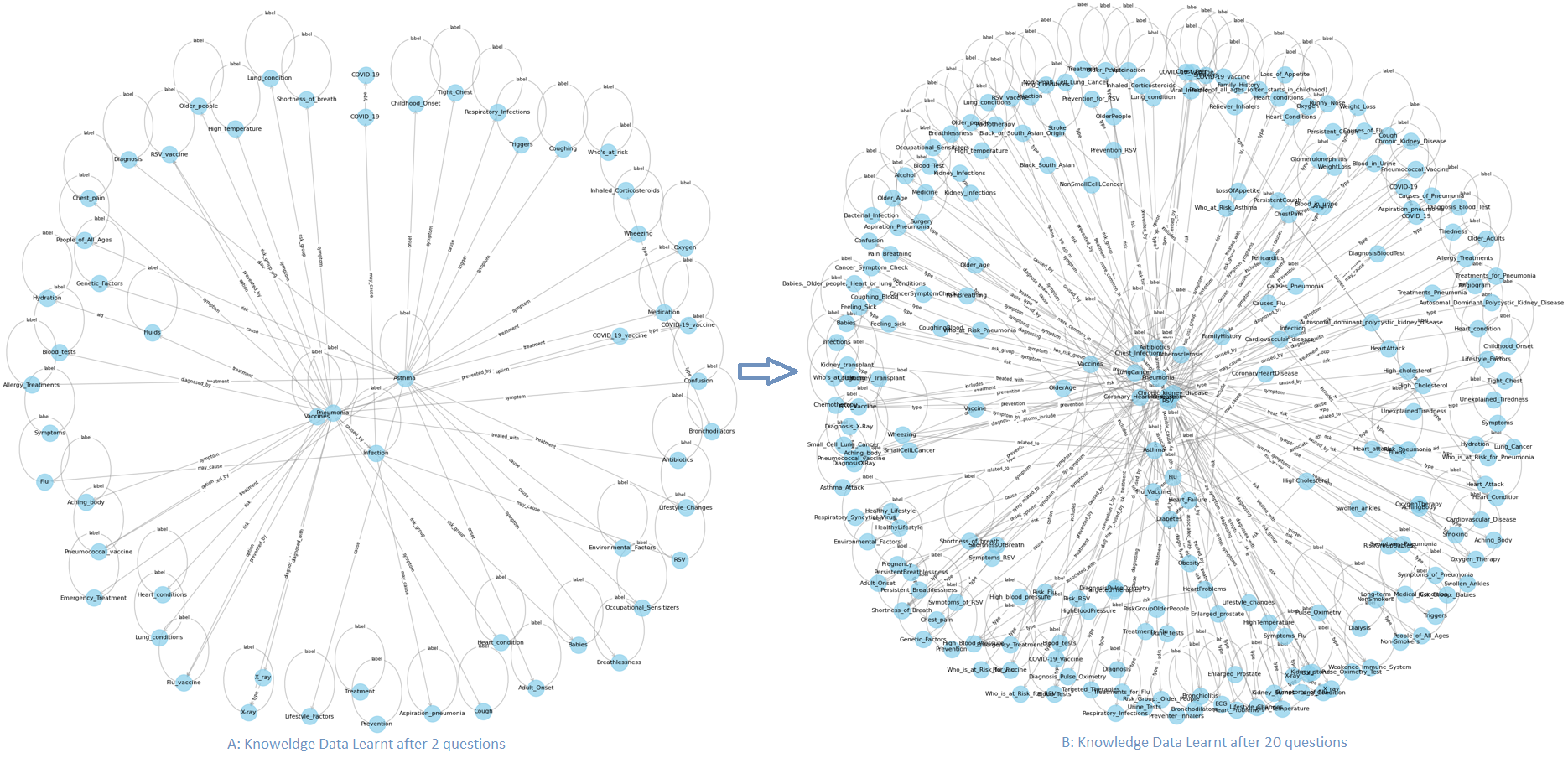}
    \caption{Knowledge Data Growing from Graph A to Graph B. The growth shows the increase in terms, triples, and unique relation types as new questions were answered.}
    \label{fig:knowledge_growth}
\end{figure*}

\subsection{Knowledge Creation and Causality Reasoning Capabilities Evaluation}

\textbf{Knowledge creation:} Figure~\ref{fig:knowledge_growth} shows the generated and growth of the Knowledge Graph during the learning process by the RAG-KG-IL Agent. The comparison from Graph A to Graph B reveals the significant differences of knowledge gained. In explicit, after only two questions and answers, the system had only learned 57 terms, 114 triples, and 19 unique relation types. The numbers are increased to 141 terms, 356 triples, and 23 unique relation types after 10 questions. This growth continued, and after 20 questions and answers, the knowledge graph had expanded to 226 terms, 420 triples, and 36 unique relation types. This indicates that the proposed system effectively learns and incorporates new information into its existing knowledge base over time, enriching the KG with valuable relationships and entities.

The results also demonstrate the advantages of adopting LightKG \cite{ref31} and LightRAG \cite{ref30} approaches. LightKG and LightRAG both propose a lightweight way of using Knowledge Graphs without having predefined ontologies but instead flexibly creating graph data and generating schemas on the fly according to ground data changes. 

\begin{figure}[h]
    \centering
    \includegraphics[width=\linewidth]{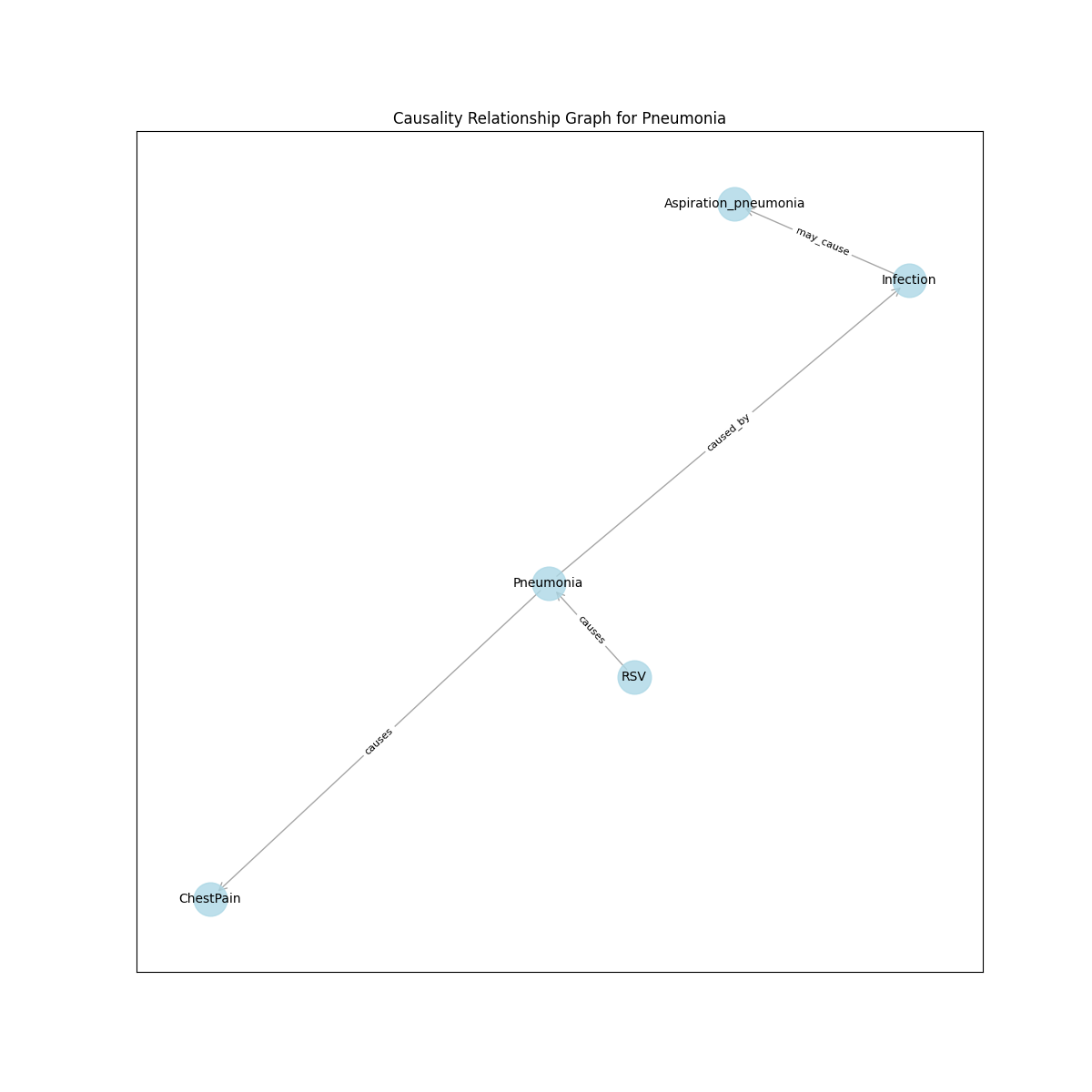}
    \caption{Knowledge reasoning capability applied to Pneumonia causes after only two questions.}
    \label{fig:reasoning_example_1}
\end{figure}
\begin{figure}[h]
    \centering
    \includegraphics[width=\linewidth]{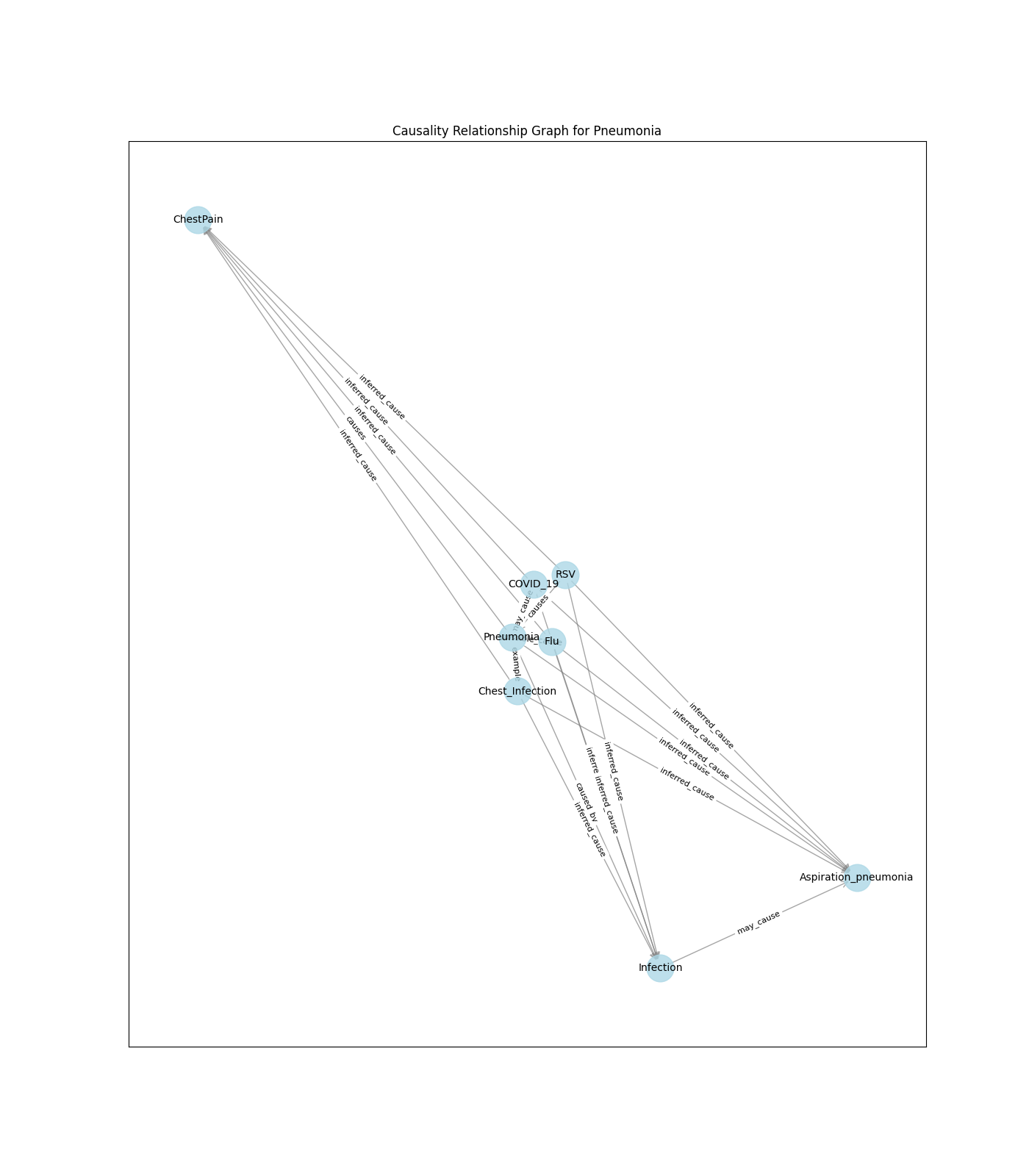}
    \caption{Knowledge reasoning capability applied to Pneumonia causes after 20 questions.}
    \label{fig:reasoning_example_2}
\end{figure}

\textbf{Causality reasoning capabilities:} Figures~\ref{fig:reasoning_example_1} and \ref{fig:reasoning_example_2} further display the system's causality reasoning capabilities by applying knowledge to the cause of Pneumonia disease. After answering only two questions, the system demonstrated an initial understanding of Pneumonia causes as Figure~\ref{fig:reasoning_example_1} demonstrates four causes relationships of three types are discovered: (1) Pneumonia - causes - chest pain and RSV; (2) Pneumonia - caused by - Infection; (3) Infection - may cause - Aspriation pneumonia. 

In contrast, After 20 questions, Figure~\ref{fig:reasoning_example_2} illustrates an enhanced understanding of the problem of Pneumonia causalities. More inferred causes relations are generated between more related diseases such as Covid-19, RSV, Flu, Pneumonia and Chest Infection as well as extra causal relationships to the Chest Pain, Aspiration Pneumonia and Infection. The enhancing understanding indicates that the system had learned and synthesized additional causes and relationships regarding Pneumonia, demonstrating its incremental reasoning capability.

\subsection{Scalability evaluation}
\begin{figure}[h]
    \centering
    \includegraphics[width=\linewidth]{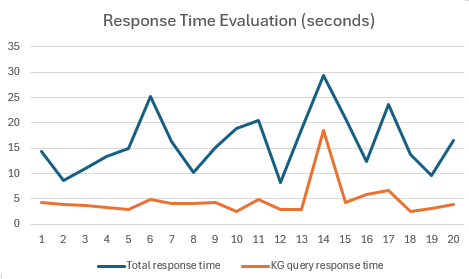}
    \caption{Scalability evaluation in terms of time cost, x: the number of questions, y: the response time in seconds}
    \label{fig:time}
\end{figure}

We recorded two types of time consumption: the total response time for answering questions and the knowledge query response (retrieval) time. The recorded data is presented in Figure~\ref{fig:time}. The data shows that the response time is discontinuous, and no significant impact was observed due to the increasing number of questions and the growing size of the knowledge graph between 1 and 20 questions. The total response time for answering questions was mainly affected by the LLM API's internet traffic, while the knowledge graph query time was primarily influenced by the size of the relevant subgraph created from the entire knowledge base with regards to a specific question. However, the evaluation is currently limited and does not include testing with large-scale knowledge graph data, as mentioned in the methodology section.    

\subsection{Accuracy Evaluation}

\begin{figure}[h]
    \centering
    \includegraphics[width=\linewidth]{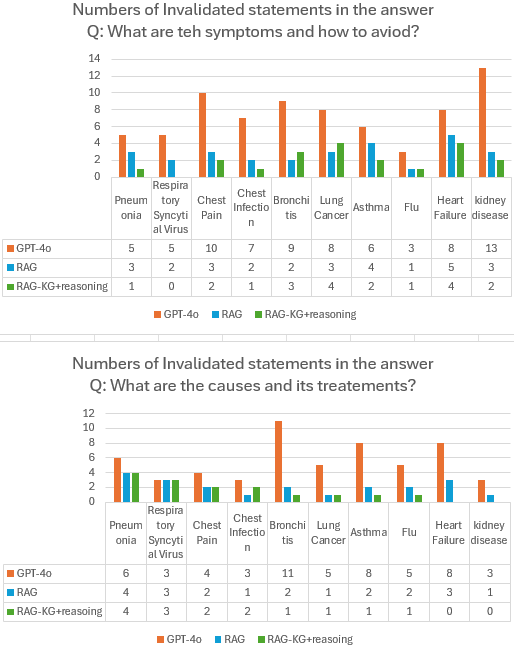}
    \caption{The numbers of invalidated statements in the answers, highlighting the relative hallucination rates of GPT-4o, RAG and RAG-KG-IL.}
    \label{fig:accuracy_comparison}
\end{figure}

Figure~\ref{fig:accuracy_comparison} illustrates the accuracy evaluation results by counting the number of hallucinated words or phrases appearing in the answers provided by the RAG-KG-IL Agent (green, bottom), RAG-only (blue, middle), and GPT-4.0 (orange, top) for two sets of 20 questions covering 10 related diseases. A lower count is preferred as it indicates fewer hallucinations. The hallucination rate refers to the presence of incorrect or misleading information not found in the ground truth data we provided. Here, "accuracy" does not denote the overall correctness but rather the alignment of the presented information with the verified ground truth data (i.e., description documents). An answer might be clinically correct but still not matching our ground truth.

In this comparison, the RAG-KG-IL Agent demonstrates a significantly reduced hallucination rate compared to both GPT-4.0 and the RAG-only system. RAG-KG-IL consistently shows the lowest hallucination count across all questions. For instance, GPT-4.0 exhibits a high hallucination count of 13 in one question, whereas RAG-KG-IL shows only 2. The total hallucination counts for RAG-KG-IL, RAG-only, and GPT-4.0 are 35, 49, and 129, respectively. This represents a substantial reduction of around 73\% in hallucinations for RAG-KG-IL compared to GPT-4.0. Additionally, RAG-only reduces hallucinations by approximately 63\% relative to GPT-4.0 but still remains significantly higher than the proposed RAG-KG-IL multi-agent framework.

\begin{figure}[h]
    \centering
    \includegraphics[width=\linewidth]{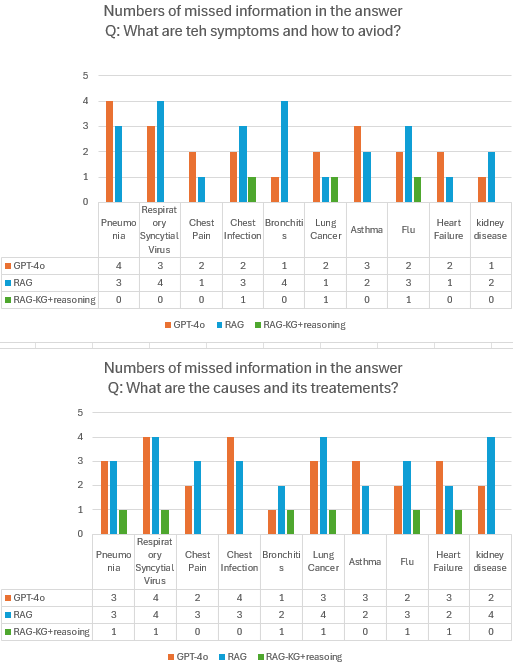}
    \caption{The numbers of missed information in the answers, highlighting the relative incompleteness rate of GPT-4o, RAG and RAG-KG-IL.}
    \label{fig:accuracy_comparison2}
\end{figure}

Furthermore, we integrated completeness evaluation with accuracy, as both the RAG-KG-IL Agent and GPT-4.0 generally provide more comprehensive information than the RAG-only framework. Figure \ref{fig:accuracy_comparison2} presents these completeness comparison results. Together with the accuracy evaluation, it is clear that while the RAG-only framework reduces hallucinations significantly, it misses essential information, leading to incomplete answers because it lacks connectivity to related documents for gathering additional relevant knowledge. GPT-4.0 covers slightly more information than RAG-only but generates significantly more hallucinations when measured against our provided ground truth.

To summarise, RAG-KG-IL has low incompleteness counts, with a maximum of 1 and often zero for many questions. In contrast, RAG-only shows the highest incompleteness, with 54 instances and a maximum count of 4 in five questions. GPT-4.0 is only marginally better than RAG-only in incompleteness, with a total of 49 instances and a maximum of 4 in two questions. Both RAG-only and GPT-4.0 miss information in nearly every single question.      

Overall, the RAG-KG-IL framework uniquely achieves low hallucination rates and high completeness in responses. This RAG-KG-IL and reasoning framework benefits from leveraging a structured knowledge source, in the form of a Knowledge Graph (KG), which verifies facts and ensures responses are grounded in accurate data. 


\textbf{Further analysis and findings:} 

\begin{figure}[h]
    \centering
    \includegraphics[width=\linewidth]{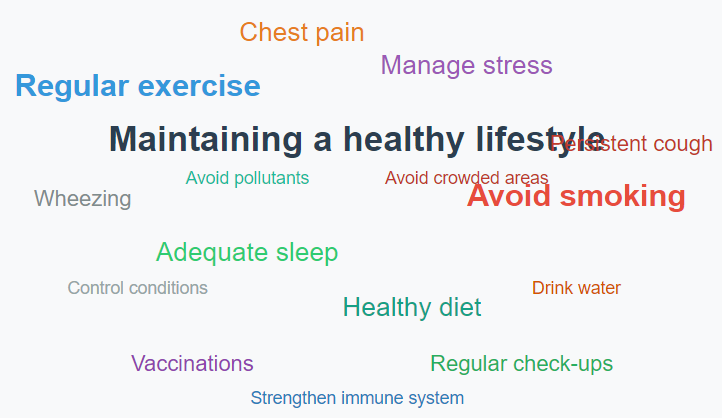}
    \caption{Top 20 frequently hallucination phrases}
    \label{fig:hrank}
\end{figure}

In examining the evaluation results, the 20 most frequent hallucination words or phrases appear in more than 63\% of answers across all three systems (see Figure~\ref{fig:hrank}). Many of these are lifestyle-related terms, likely influenced by the pre-training data of GPT-4.0, indicating a fundamental limitation of its statistical word prediction approach.

\section{Conclusion and future work}

This paper presented an integrated framework combining Retrieval-Augmented Generation (RAG) and Knowledge Graph (KG) reasoning techniques to enhance the accuracy and knowledge evolution capabilities of generative AI models. The proposed RAG-KG-IL framework effectively reduces hallucination by integrating structured knowledge and performing reasoning based on learned information.

The evaluation results demonstrate that our framework achieves significant improvements in reducing hallucination rates compared to traditional models, such as GPT-4o, by incorporating structured knowledge. The RAG-KG-IL demonstrated a substantial reduction in hallucination (73 percentage lower compared to GPT-4o) and significant knowledge growth, as evidenced by the expansion of its knowledge graph after answering multiple questions. The system's reasoning capabilities also improved over time, illustrating the ability to generate richer and more accurate relationships between entities within the knowledge graph. Furthermore, the adoption of lightweight approaches like LightKG and LightRAG enabled the flexible and incremental growth of the knowledge base, enhancing adaptability and real-time knowledge integration.

To further improve the capabilities of the RAG-KG-IL, several avenues for future work are proposed:

Multi-Modal Data Integration: Future research will focus on extending the current framework to integrate multi-modal data sources.

Advanced Security and Privacy Mechanisms: Given the sensitive nature of domains like healthcare, future work will explore incorporating advanced security protocols and privacy-preserving techniques to ensure that the knowledge graph and the retrieved information are both reliable and secure.


Scalability and Distributed Architectures: Future work will involve enhancing the scalability of the framework by exploring distributed architectures that can support large-scale knowledge graphs and handle high query loads in real-time applications.

\end{document}